%% file: main.tex
\documentclass[runningheads]{llncs}

\usepackage[T1]{fontenc}
\usepackage{graphicx}
\usepackage{float}
\usepackage{hyperref}
\usepackage{adjustbox}
\input{macro.tex}

\begin{document}
\title{Rule2Text: Natural Language Explanation of Logical Rules in Knowledge Graphs}

\titlerunning{Rule2Text: NL Explanation of Logical Rules in KGs} 

\author{Nasim Shirvani-Mahdavi\orcidID{
0009-0006-2733-2242} \and Devin Wingfield \and Amin Ghasemi \and Chengkai Li\orcidID{0000-0002-1724-8278}}
\authorrunning{N. Shirvani-Mahdavi et al.}
\institute{University of Texas at Arlington,
  701 S Nedderman Dr, Arlington, TX, 76019, USA
  \email{\{nasim.shirvanimahdavi2,dtw8917,mxg6185\}@mavs.uta.edu}, cli@uta.edu}

\maketitle

\input{sec-abstract}

\input{sec-introduction}

\input{sec-rule}

\input{sec-methodology}

\input{sec-results}
\input{sec-conclusion}

%\bibliographystyle{splncs04}
%\bibliography{refs}    

\input{main.bbl}
\end{document}

%% file: macro.tex
\newcommand{\entity}[1]{\texttt{\small\allowbreak #1}}
\newcommand{\rel}[1]{\textsl{\small #1}}
\newcommand{\triple}[3]{(\entity{#1}, \rel{#2}, \entity{#3})}

%% file: sec-abstract.tex
\begin{abstract}
Knowledge graphs (KGs) often contain sufficient information to support the inference of new facts. Identifying logical rules not only improves the completeness of a knowledge graph but also enables the detection of potential errors, reveals subtle data patterns, and enhances the overall capacity for reasoning and interpretation. However, the complexity of such rules, combined with the unique labeling conventions of each KG, can make them difficult for humans to understand. In this paper, we explore the potential of large language models to generate natural language explanations for logical rules. Specifically, we extract logical rules using the AMIE 3.5.1 rule discovery algorithm from the benchmark dataset FB15k-237 and two large-scale datasets, FB-CVT-REV and FB+CVT-REV. We examine various prompting strategies, including zero- and few-shot prompting, including variable entity types, and chain-of-thought reasoning. We conduct a comprehensive human evaluation of the generated explanations based on correctness, clarity, and hallucination, and also assess the use of large language models as automatic judges. Our results demonstrate promising performance in terms of explanation correctness and clarity, although several challenges remain for future research. All scripts and data used in this study are publicly available at \href{https://github.com/idirlab/KGRule2NL}{https://github.com/idirlab/KGRule2NL}.

\end{abstract}

%% file: sec-introduction.tex
\vspace{-0.9cm}\section{Introduction}\label{sec:intro}
\vspace{-0.4cm}
Knowledge graphs (KGs) encode factual information as triples of the form (subject \entity{s}, predicate \rel{p}, object \entity{o}). They are integral to a wide range of artificial intelligence tasks and applications~\cite{ji2021survey,kiafar2025mena}. Although large-scale KGs (e.g., Freebase~\cite{bollacker2008freebase} and Wikidata~\cite{vrandevcic2014wikidata}) contain a vast number of triples, they are often incomplete, which adversely affects their usefulness in downstream applications. However, KGs often hold sufficient information to infer new facts~\cite{amieplus,shirvani2025large}. 
For example, if a KG indicates that a certain woman is the mother of a child, it is quite likely that her husband is the child’s father. 
Identifying such rules can help infer highly probable missing facts which can be further verified by human data workers or experts. 
In addition to enhancing the completeness of KGs, such rules can also aid in detecting potential errors, deepening our understanding of the data's inherent patterns, and facilitating reasoning and interpretability~\cite{nakashole2012query,amieplus}. 
Rule learning systems, such as  AMIE~\cite{galarraga2013amie,betz2023pyclause} and AnyBURL~\cite{meilicke2024anytime}, derive Horn rules for symbolic reasoning and link prediction in KGs. These rules can serve as explanations for specific predictions; for instance, such rules can assist domain scientists in uncovering underlying missing relationships within their data.

\vspace{-0.1cm}However, rules are often challenging to comprehend for humans, especially for non-experts. The difficulty arises from the abstract logical structure and the complexity of the rules; the number of logical components, referred to as atoms, as well as the nuanced nature of entity and relation labels within each KG. For instance, as explained \sloppy in~\cite{shirvani2023comprehensive}, label of predicates in the Freebase dataset follow the format /[domain]/[type]/[label] (e.g., 
\rel{/american\_football/player\_rushing\_statistics/team}). Without proper background knowledge about such differences in KG labels, evaluating logical rules can become cumbersome.

\vspace{-0.1cm}One way to address this challenge is by providing natural language explanations for logical rules, which enhance accessibility and usability, aid KG management in cross-disciplinary contexts, and improve transparency for researchers and practitioners. Pre-defined templates can generate such explanations, but this approach is not scalable, as it is impractical to manually extract all logical rules from a large KG and define templates for each. To handle unseen rules, solutions leveraging large language models (LLMs) are promising due to LLMs' generative abilities and generalization capability. Related work has focused on natural language generation from logical forms~\cite{wu2023generating,chen2020logic2text}, natural language generation from KGs~\cite{shi2023hallucination}, encoding and translating natural rules~\cite{clark2020transformers,aesoy2023rule}, and rule-based reasoning with LLMs~\cite{servantez2024chain,yang2023logical}.

\vspace{-0.1cm}To the best of our knowledge, this is the first work to examine the effectiveness of LLMs in generating natural language explanations for logical rules. We mined the rules by the AMIE 3.5.1 algorithm, the latest version released in 2024, using the widely used cross-domain benchmark dataset FB15k-237~\cite{bordes2013translating} and two properly preprocessed large-scale variants of the Freebase dataset, FB-CVT-REV and FB+CVT-REV~\cite{shirvani2023comprehensive} (Section~\ref{sec:rules}). We investigated a range of prompting strategies, such as zero and few-shot prompting~\cite{kermani2025systematic}, incorporating an instance of the rule, including variable entity types and Chain-of-Thought (CoT) reasoning~\cite{wei2022chain}  (Section~\ref{sec:methodology}). To evaluate the quality of the generated explanations, we conducted detailed human evaluations based on criteria such as correctness, clarity, and hallucination.  Additionally, we explored the potential of LLM-as-a-judge~\cite{zheng2023judging} for this task (Section~\ref{sec:results}). Our findings indicate that combining CoT prompting with variable type information yields the most accurate and readable explanations. Overall, our findings highlight a promising direction for this task. We conclude the work and outline potential avenues for future research in Section~\ref{sec:conclusion}. All the scripts and data produced from this work are available from our GitHub repository at \href{https://github.com/idirlab/KGRule2NL}{https://github.com/idirlab/KGRule2NL}.

%% file: sec-rule.tex
\vspace{-0.4cm}\section{Rule Selection from Knowledge Graphs}\label{sec:rules}\vspace{-0.2cm}
\subsection{Rule Mining Algorithm}\vspace{-0.2cm}
We employed AMIE 3.5.1, a well-established rule learning system in its latest version, due to its comprehensive metrics for rule evaluation as well as its proven compatibility with our chosen benchmark datasets. In AMIE, a rule has a body (antecedent) and a head (consequent), represented as $B_1 \land B_2 \land \ldots \land B_n \Rightarrow H$
, or in simplified form \(\overrightarrow{B} \Rightarrow H\).  
The body consists of multiple \textit{atoms} $B_1$, $\ldots$, $B_n$ and the head $H$ itself is also an atom. In an atom \rel{r}(\entity{h},\entity{t}), which is another representation of a triple \triple{h}{r}{t}, the subject and/or the object are variables to be instantiated. The prediction of the head can be carried out when all the body atoms can be instantiated in the KG. In AMIE, the concept of \textit{support} quantifies the amount of evidence (i.e., correct predictions) for each rule in the data. It is defined as the number of distinct (subject, object) pairs in the head of all valid instantiations of the rule in the KG. The concept of \textit{head coverage}, a proportional version of \textit{support},  is the fraction of \textit{support} over the number of facts in relation \rel{r}, where \rel{r} is the relation in the head atom. The \textit{standard confidence} of a rule is the fraction of \textit{support} over the number of instantiations of the rule body. To mine the rules, we used the default settings of AMIE for optimized performance, with minimum thresholds of 0.1 for \textit{head coverage} and \textit{standard confidence}, and a maximum threshold of 3 for the number of atoms.

\vspace{-0.5 cm}\subsection{Datasets}\vspace{-0.2 cm}
For our experiments, we leveraged three datasets. FB15k-237, a small subset of the Freebase dataset, was selected as it is a widely used benchmark for KG completion, recognized for avoiding the data leakage issues of FB15k~\cite{bordes2013translating}. Its multi-domain coverage makes it well-suited for extracting logical rules with diverse relations. FB-CVT-REV and FB+CVT-REV~\cite{shirvani2023comprehensive} datasets (Statistics shown in Table~\ref{tab:datastat}) are large-scale variants of the Freebase dataset designed to eliminate the data leakage issue previously identified in FB15k. FB+CVT-REV includes mediator entities (i.e., Compound Value Type nodes) originally present in Freebase to represent n-ary relations. In contrast, FB-CVT-REV converts n-ary relationships centered on a CVT node into binary relations by concatenating the edges that connect entities through the CVT node, a method also used in FB15k-237. As shown in Table~\ref{tab:datastat}, the conversion process has resulted in a higher number of rules in these two datasets compared to those in FB+CVT-REV. Including these datasets facilitates the analysis of large-scale data and the effects of mediator nodes and concatenated relationships on the derived rules and generated explanations.

The label of a concatenated relation is formed by merging the labels of two underlying relations. As a result, the label becomes lengthy, taking the format of \rel{domain1/type1/label1./domain2/type2/label2}. Notably, the domains and even types can sometimes be identical in concatenated labels, but label1 and label2 are always distinct. This format differs from the simpler structure of standard relations, which follow the format of \rel{domain/type/label}. Thus, this added complexity can pose a greater challenge for LLMs in generating natural language explanations.

%% file: sec-methodology.tex
\vspace{-0.4 cm}\section{Methodology}\label{sec:methodology}
\vspace{-0.2 cm}
\subsection{Prompting Strategies}
To generate natural language explanations for logical rules, we conducted our experiments in three phases. All scripts and prompts, rules, generated explanations, and annotated data are available at our GitHub repository. In all prompts used in our experiments, we provided background knowledge to the models to enhance their understanding of the syntax and labels of the datasets. This content includes the format of predicates in the datasets as mentioned in Section~\ref{sec:rules}. This background knowledge is particularly important because, in rules that involve concatenated relations, the resulting lengthy labels with multiple components can easily confuse the model.

\vspace{-0.2 cm}\textbf{Phase 1: Zero-Shot vs. Few-Shot Prompting}\hspace{3 mm} In the first phase, we compared zero-shot and few-shot prompting strategies using rules from a small subset of the Freebase dataset, specifically FB15k-237. The objective was to assess the impact of in-context examples on explanation quality and establish a baseline. The few-shot prompt includes two (rule, explanation) pairs as examples. In this phase and phase 2, we employed OpenAI’s GPT-3.5 Turbo model for its balance of performance, efficiency, and cost-effectiveness. A total of 100 rules with the highest head coverage were selected for human evaluation, covering a broad range of domains, from music and media to medicine and space. To ensure the quality of the annotations and mitigate potential subjectivity, we tasked three individuals with annotating the data. For each rule, annotators were shown both the rule and a concrete instantiation to aid understanding, along with two generated explanations, one from zero-shot prompting and one from few-shot. Their task was to identify which explanation better captured the semantics of the rule. In cases of comparable semantic accuracy, the more naturally worded explanation was preferred. After selecting the better explanation, annotators rated it using the evaluation metrics described in Section~\ref{sec:eval}. As discussed in Section~\ref{sec:results}, the few-shot prompting strategy did not yield significant improvements over the zero-shot baseline.

\vspace{-0.2 cm}\textbf{Phase 2: Utilizing Variable Entity Types in The Prompt}\hspace{3 mm} This phase initially incorporated rule instantiations into the prompt design. However, analysis of the generated explanations revealed persistent limitations in the model's ability to identify variable entity types, leading us to adopt integration of these types in the prompt. For instance, in the rule \entity{?b} \rel{/time/event/instance\_of\_recurring\_event} \entity{World Series} => \entity{World Series} \rel{/sports/sports\_championship/events} \entity{?b}, \entity{World Series} is a constant entity and \entity{?b} is a variable entity. In Freebase datasets, entities can belong to multiple types. Consequently, each variable entity is associated with a list of types. Given an edge type and its edge instances, there is \emph{almost} a function that maps from the edge type to a type that all subjects in the edge instances belong to, and similarly, \emph{almost} such a function for objects\cite{shirvani2023comprehensive}. For the example above, the variable \entity{?b}'s types are either \entity{/time/event} or \entity{/sports/sports\_championship\_event}.
For this phase, three annotators annotated 100 rules, rules with the highest head coverage, 50 top rules from FB-CVT-REV, and 50 from FB+CVT-REV.  Unlike the previous phase, the annotators were asked to complete metric evaluations for explanations from both prompts, the zero-shot prompt as our baseline, and the prompt incorporating variable type. As discussed in Section~\ref{sec:results}, our findings show that providing variable type information significantly improved the model's performance in generating accurate explanations.

\textbf{Phase 3: Comparing Models \& Chain-of-Thought Prompting}\hspace{3 mm} Building on the strong impact of incorporating variable entity types into the prompts, we further leveraged the reasoning capabilities enabled by CoT prompting. This prompt guides the model through five reasoning steps. First, it parses the rule and identifies its components, including constant entities, variable entities, and relations. Second, for each variable entity, it determines the most contextually relevant type. Third, it interprets each atom in the rule using the selected types. Fourth, it synthesizes the information to infer the rule’s overall implication. Finally, it generates a concise natural language explanation. The prompt also includes two illustrative examples with CoT reasoning to support the model’s understanding. In this phase, we expanded our evaluation to include two additional models, GPT-4o Mini and Gemini 2.0 Flash, alongside GPT-3.5 Turbo. These models were selected to provide a balanced comparison in terms of performance, efficiency, and cost-effectiveness. Three annotators evaluated new explanations, generated via CoT prompting by the three models, for the same set of rules used in phase 2. As discussed in Section~\ref{sec:results}, GPT-3.5 Turbo shows improved performance compared to phase 2, while Gemini 2.0 Flash achieves the highest overall performance, followed by GPT-4o Mini. 
     
\vspace{-0.4 cm}\subsection{Evaluation Metrics for Generated Explanations}\label{sec:eval} \vspace{-0.3 cm}
To evaluate the generated explanations, we used the following metrics for human and automatic evaluation.

\textit{Correctness}: Evaluation of the explanation's accuracy on a scale from 1 (completely incorrect) to 5 (fully correct). Correctness refers to the explanation's inclusion of all components of the rule, presented in the exact logical order specified by the rule.

\textit{Clarity}: Evaluation of the explanation on a scale from 1 (very unclear) to 5 (very clear). Clarity refers to the ease with which the explanation can be understood and how naturally it reads. This metric exclusively assesses the explanation, independent of the correctness of the underlying rule.

\textit{\# of missed entities}: 
The number of entities present in the rule but not stated in the explanation.

\textit{\# of missed relations}: 
The number of relations (i.e., predicates) present in the rule but not stated in the explanation.

\textit{\# of hallucinated entities}:
The number of entities absent from the rule but incorrectly stated in the explanation.

\textit{\# of hallucinated relations}: 
The number of relations absent from the rule but incorrectly stated in the explanation.

\textit{Rule logicalness}: Although the meaningfulness of a rule is not directly related to the generation of explanations, we asked the annotators to rate the rules on a scale from 1 (not logically sound) to 2 (moderately logical), and 3 (logically sound). This metric exclusively evaluates the rule itself, without considering the explanation. 

\textit{Perplexity}: Given the absence of reference sentences for comparison with the explanations, as our automatic evaluation metric, we computed perplexity using GPT-2. While it is a useful measure of the model's fluency and coherence, it is not an indication of the correctness of the explanations.

%% file: sec-results.tex
\vspace{-0.4cm}\section{Experiments \& Results}\label{sec:results}
\vspace{-0.3cm}
\textbf{Phase 1} \hspace{0.1cm}The annotated data, available on our GitHub repository, represents an aggregation of input from three annotators. For each rule, we select the explanation receiving the majority vote and calculate the average of the measures for that chosen explanation only. For instance, if annotators 1 and 2 selected the explanation generated from the zero-shot prompt for a particular rule, while the third annotator chose the explanation from the few-shot prompt, we only averaged the measures provided by annotators 1 and 2 for that rule.  

Table~\ref{table:results} presents the average of all measures for all annotated rules (denoted as all), as well as for explanations generated from zero and few-shot prompts separately (denoted as prompt 1 and prompt 2, respectively). Specifically, for rules where the explanation generated from prompt 1 was favored by the majority of annotators, the table reports the average measures for those explanations. The same approach is applied to rules where explanations from prompt 2 were preferred. Additionally, we provide these measures for the subset of rules where annotators unanimously selected the same explanation (denoted as unanimous), comparing them to the remaining rules with majority voting (denoted as majority). The measures \# missed entities, \# missed relations, \# hallucinated entities, \# hallucinated relations, correctness, clarity, logicalness, and perplexity are denoted as m ent, m rel, h ent, h rel, correctness, clarity, logical, and perplexity in Table~\ref{table:results} (and ~\ref{table:results2}) respectively.

These results demonstrate the model, overall, generates relatively accurate and clear explanations with low perplexity. Among the 100 sentences selected for human annotation, 49 were assigned to explanation 2, derived from the few-shot prompt, while the remaining sentences were assigned to explanation 1. Notably, annotators reached unanimous agreement on 48\% of the rules. Furthermore, the number of missed or hallucinated elements is negligible. Our observations indicate that most hallucinations stem from the labels of relations, particularly concatenated relations. The model tends to generate additional entities or relations to explain the complex labels associated with concatenated relations. 

\begin{table*}
\centering 
\begin{minipage}{0.28\textwidth}
\setlength{\tabcolsep}{2pt}
\vspace{-0.5 cm}
\caption{Statistics of the datasets}\label{tab:datastat}
\vspace{-0.3 cm}
\scalebox{0.63}{\begin{tabular}{l|cc}
\hline
\textbf{dataset}    & \textbf{\# of triples} & \textbf{\# of rules} \\ \hline
FB15k-237 & 310,116 & 6,320 \\ 
FB-CVT-REV & 125,124,274 & 14,355  \\
FB+CVT-REV & 134,213,735 & 2,965  \\ \hline
\end{tabular}
}\vspace{-0.7 cm}
\end{minipage} 
\hfill  
\begin{minipage}{0.67\textwidth}         
\setlength{\tabcolsep}{1.3pt}
\vspace{-0.5 cm}
\caption{Evaluation results on the annotated data in phase 1}\label{table:results}
\vspace{-0.3 cm}
\centering
\scalebox{0.65}{\begin{tabular}{l|cccc|ccc|c}
\hline
                 
\textbf{}  & \textbf{m ent$^\downarrow$}    & \textbf{m rel$^\downarrow$} & \textbf{h ent$^\downarrow$} & \textbf{h rel$^\downarrow$}& \textbf{correctness$^\uparrow$} & \textbf{clarity$^\uparrow$} & \textbf{logical$^\uparrow$} & \textbf{perplexity$^\downarrow$} \\ \hline
\textbf{all}  & 0.10   & 0.04   & 0.29    & 0.10   & 4.36   & 4.67 &2.36 & 36.14   \\ 
\textbf{prompt 1} & 0.14   & 0.05   & 0.25    & 0.07   & 4.40   & 4.69 & 2.29   & 37.85   \\ 
\textbf{prompt 2}  & 0.06   & 0.03   & 0.34    & 0.12   & 4.32   & 4.64  & 2.44   & 34.36  \\ 
\textbf{unanimous}   & 0.13   & 0.03   & 0.35    & 0.12  & 4.34   & 4.68  & 2.29   & 33.80  \\ 
\textbf{majority}   & 0.08   & 0.05  & 0.24    & 0.07   & 4.37   & 4.66  & 2.43   & 38.30  \\ 

\hline
\end{tabular}}
\vspace{-1 cm}
        \end{minipage}\vspace{3mm} 
\end{table*}
\textbf{Phase 2} \hspace{0.1cm} Table~\ref{table:results2} presents the results for this phase, averaged across all annotators. Explanation 2, generated using the variable type prompt, consistently shows higher correctness and clarity across all categories, highlighting the importance of type information for model comprehension. Both explanation types have minimal missing entities and relations. However, Explanation 2 also shows slightly higher hallucination rates and increased perplexity. Rules with three atoms and those involving concatenated relations generally receive lower correctness and clarity scores, likely due to their increased complexity. Interestingly, despite these lower scores, annotators rated the rules from these two categories as more logically coherent.

\begin{table*}[ht]
\centering
\setlength{\tabcolsep}{1.3pt}

\caption{Evaluation results on the annotated data in phase 2}\label{table:results2}
\vspace{-0.3 cm}
\scalebox{0.57}{
\begin{tabular}{l|c|cccc|cc|c|cccc|cc|c}
\hline
& & \multicolumn{7}{c|}{\textbf{explanation from zero-shot prompt}} & \multicolumn{7}{c}{\textbf{explanation from variable type prompt}} \\
\hline
\textbf{}  & \textbf{logical$^\uparrow$} & \textbf{m ent$^\downarrow$} & \textbf{m rel$^\downarrow$} & \textbf{h ent$^\downarrow$} & \textbf{h rel$^\downarrow$} & \textbf{correct$^\uparrow$} & \textbf{clarity$^\uparrow$} &  \textbf{perplexity$^\downarrow$} & \textbf{m ent$^\downarrow$} & \textbf{m rel$^\downarrow$} & \textbf{h ent$^\downarrow$} & \textbf{h rel$^\downarrow$} & \textbf{correct$^\uparrow$} & \textbf{clarity$^\uparrow$} & \textbf{perplexity$^\downarrow$}  \\ 

\hline
\textbf{all}             & 2.58&	0.06&	0.10&	0.22	&0.09	&3.94	&4.12&29.05&	0.05&	0.07&	0.21	&0.13 &	4.21&	4.19 & 33.07  \\ 
\textbf{2 atoms}        & 2.50 & 0.03 &0.04 &0.08 &0.05& 4.22 &4.35& 34.10 &0.31 &0.41& 0.15 &0.16 &4.25 &4.30  & 38.59 \\ 
\textbf{3 atoms}        & 2.62 & 0.08 &0.13 & 0.31 & 0.12 &3.78 &3.99& 26.21& 0.07& 0.08& 0.24& 0.11& 4.18 &4.12  & 29.97 \\ 
\textbf{binary}     & 2.59 & 0.08 & 0.10 & 0.18 & 0.08 & 4.04 & 4.22 & 31.02 & 0.06& 0.03& 0.20 &0.11& 4.32& 4.28   & 34.11 \\ 
\textbf{mediator}        & 2.51 & 0.08 &0.13& 0.16 &0.06& 4.15 &4.13& 24.22& 0.01& 0.11& 0.16 &0.06& 4.36& 4.2  & 28.65 \\ 
\textbf{concatenated}       & 2.60 & 0.02 & 0.08 & 0.35 & 0.15 & 3.63 & 3.91 & 27.63 & 0.05 & 0.11 & 0.25 &0.20 &3.88 &3.99  & 33.33  \\ 
\hline
\end{tabular}}
\vspace{-0.7 cm}
\end{table*}

\textbf{Phase 3} \hspace{0.1cm} Given the negligible number of hallucinated and missing entities and relations, we evaluated the explanations in phase 3 using only correctness, clarity, and perplexity. Table~\ref{table:results3} presents the results. Overall, the models exhibit trends similar to those observed in Phase 2. For example, all models perform better on shorter rules, particularly those with only two atoms, and achieve higher performance on rules involving only binary relations compared to those with concatenated ones. GPT-3.5 Turbo shows improved performance with CoT prompting compared to its performance using only variable entities. This improvement is consistent across all categories except for rules that include mediator nodes. GPT-4o Mini is the second-best performing model and demonstrates relatively strong performance on rules containing at least one concatenated relation. Gemini 2.0 Flash demonstrates the best overall performance. Its explanations are the most concise, though in rare instances, it includes remarks such as, “Note: This rule is likely flawed.” Notably, the lowest clarity scores across all models are observed for rules involving mediator nodes. Additionally, most models exhibit their highest perplexity on rules with only two atoms, which is somewhat unexpected given the simplicity of these rules.

\begin{table*}[ht]
\centering
\setlength{\tabcolsep}{1.3pt}
\vspace{-0.7 cm}
\caption{Evaluation results on the annotated data in phase 3}\label{table:results3}
\vspace{-0.2 cm}
\scalebox{0.7}{
\begin{tabular}{l|cc|c|cc|c|cc|c}
\hline
& \multicolumn{3}{c|}{\textbf{GPT-3.5 Turbo}} & \multicolumn{3}{c|}{\textbf{GPT-4o mini}} & \multicolumn{3}{c}{\textbf{Gemini 2.0 Flash}}\\
\hline
\textbf{}   & \textbf{correct$^\uparrow$} & \textbf{clarity$^\uparrow$} &  \textbf{perplexity$^\downarrow$}  & \textbf{correct$^\uparrow$} & \textbf{clarity$^\uparrow$} & \textbf{perplexity$^\downarrow$} & \textbf{correct$^\uparrow$} & \textbf{clarity$^\uparrow$} & \textbf{perplexity$^\downarrow$} \\ 

\hline
\textbf{all}             & 4.28&	4.26&	32.40&	4.45	&4.53	&31.57	&4.67&4.70&	27.19	 \\ 
\textbf{2 atoms}        & 4.38 & 4.43 &34.08 &4.52 &4.62& 40.96 &4.80& 4.76 &29.98  \\ 
\textbf{3 atoms}        & 4.22 & 4.17 &31.46 & 4.42 & 4.51 &26.26 &4.61& 4.68& 25.62 \\ 
\textbf{binary}     & 4.40 & 4.42 & 34.58 & 4.50 & 4.58 & 33.52 & 4.70 & 4.71 & 27.77 \\ 
\textbf{mediator}        & 4.13 & 4.07 &26.26& 4.24 &4.49& 26.82 &4.69& 4.63& 26.92 \\ 
\textbf{concatenated}       & 4.10 & 4.07 & 31.57 & 4.50 & 4.51 & 30.38 & 4.63 & 4.75 & 26.19\\ 
\hline
\end{tabular}}
\vspace{-0.7 cm}
\end{table*}
\textbf{LLM-as-a-Judge} \hspace{0.1cm}
One of the limitations of this work is the absence of ground truth data, which restricts our ability to fine-tune models effectively. A potential solution lies in leveraging the LLM-as-a-judge approach. If a reliably fair and consistent judge model can be designed, it becomes possible to use a strong model, such as Gemini 2.0 Flash, to generate (rule, explanation) pairs. The judge can then evaluate these pairs, and those receiving high scores can be treated as pseudo-ground truth for fine-tuning smaller open-source models. Additionally, low-scoring examples can be analyzed to identify patterns and better understand the types of explanations or rules that pose challenges for the model. 

To explore this direction, we developed an LLM-as-a-judge prompt. LLM-based judges often exhibit bias toward models from their family~\cite{panickssery2024llm}, for example, GPT models tend to favor responses generated by other GPT models. To account for this potential bias, we evaluated the performance of the two best models, GPT-4o Mini and Gemini 2.0 Flash, using both GPT-4o Mini and Gemini 2.0 Flash as judges. This resulted in a total of four evaluation settings for a more balanced comparison. Since clarity can be a highly subjective metric, we focused our analysis on correctness. The information provided to the LLM judges was identical to that given to human annotators: the rule, an instance of the rule, the list of variable types, and the explanation. Table~\ref{table:correlation} presents the correlation between LLM judges and human annotators. Because annotator scores are averaged across multiple individuals, they are represented as floating-point numbers, whereas LLM judge scores are integers. To ensure a fair comparison, we rounded the annotator scores to the nearest whole number before computing correlation coefficients. We used Spearman correlation to measure rank-order agreement, assessing how similarly the judges and annotators rank the explanations. Pearson correlation was used to evaluate the strength of the linear relationship between their actual scores. Both LLM judges show moderate agreement with annotators on the correctness of explanations generated by GPT-4o Mini. Gemini 2.0 Flash also aligned reasonably well with annotators when evaluating its own outputs, whereas GPT-4o Mini showed weak agreement in that setting. Although these results are not ideal, they point to a promising direction for future work in leveraging LLMs for scalable evaluation and dataset generation.

\begin{table}[h]
\centering
\vspace{-0.5 cm}
\caption{Correlation between LLM judges and annotators for correctness}
\vspace{-0.2 cm}
\label{table:correlation}
\scalebox{0.73}{\begin{tabular}{l|l|c|c}
\hline
\textbf{Judge} & \textbf{Explanation generated by} & \textbf{Spearman} & \textbf{Pearson} \\
\hline
GPT-4o mini & GPT-4o mini & 0.528 & 0.595 \\
Gemini 2.0 Flash & GPT-4o mini & 0.498 & 0.603 \\
GPT-4o mini & Gemini 2.0 Flash & 0.221 & 0.208 \\
Gemini 2.0 Flash & Gemini 2.0 Flash & 0.429 & 0.527 \\
\hline
\end{tabular}}
\vspace{-1.2 cm}
\end{table}

%% file: sec-conclusion.tex
\section{Conclusion \& Future Work}\label{sec:conclusion}
\vspace{-2mm}

We employed three LLMs with multiple prompting strategies to generate natural language explanations for logical rules extracted by the AMIE algorithm from three datasets at varying scales. Human evaluation indicated encouraging results regarding accuracy and clarity, although rule complexity presents challenges for future research. Our findings indicate that the combination of Chain-of-Thought prompting and variable type information yields the most accurate and readable explanations. Future research can extend this work by evaluating LLM performance on more complex rules beyond AMIE’s extraction capabilities, exploring additional knowledge bases such as Wikidata, which encode facts differently, and constructing reference explanations to fine-tune LLMs for improved generation quality.
\vspace{-0.3 cm}
\begin{credits}
\subsubsection{\ackname} This material is based upon work supported by the National Science Foundation under Grants 
TIP-2333834. We also extend our gratitude to the Texas Advanced Computing Center (TACC) for providing computing resources for this work’s experimentation.
\end{credits}